\documentclass[letterpaper, 10 pt, conference]{ieeeconf}  
\usepackage{amsmath, amssymb}
\usepackage{algorithm}
\usepackage{algpseudocode}
\usepackage{amsmath}
\usepackage{afterpage}
\usepackage{graphicx}

\IEEEoverridecommandlockouts                              

\usepackage{todonotes}
\usepackage{booktabs}
\usepackage{multirow} 
\usepackage{array}    
\usepackage{url}
\usepackage{graphicx}
\usepackage{bbm}
\newtheorem{definition}{Definition}
\usepackage[export]{adjustbox}
\usepackage{placeins}

\usepackage{caption}
\usepackage{subcaption}
\title{\LARGE \bf
Safe Human Robot Navigation in Warehouse Scenario
}

\author{Seth Farrell*$^{1}$, Chenghao Li*$^{1}$, Hongzhan Yu$^{1}$, Ryo Yoshimitsu$^{2}$, Sicun Gao$^{1}$ and Henrik I. Christensen$^{1}$
\thanks{*These members contributed equally to this publication.}
\thanks{This work was supported by IHI Corporation}
\thanks{Affiliation: 1.UCSD, 2.IHI Japan. Email:  
        {\tt\small swfarrel@ucsd.edu}}%
}

\begin{document}

\maketitle
\thispagestyle{empty}
\pagestyle{empty}

\begin{abstract}
The integration of autonomous mobile robots (AMRs) in industrial environments, particularly warehouses, has revolutionized logistics and operational efficiency. However, ensuring the safety of human workers in dynamic, shared spaces remains a critical challenge. This work proposes a novel methodology that leverages control barrier functions (CBFs) to enhance safety in warehouse navigation. By integrating learning-based CBFs with the Open Robotics Middleware Framework (OpenRMF), the system achieves adaptive and safety-enhanced controls in multi-robot, multi-agent scenarios. Experiments conducted using various robot platforms demonstrate the efficacy of the proposed approach in avoiding static and dynamic obstacles, including human pedestrians. Our experiments evaluate different scenarios in which the number of robots, robot platforms, speed, and number of obstacles are varied, from which we achieve promising performance.
\end{abstract}

\section{INTRODUCTION}

In recent decades, the industrial sector, particularly warehouse operations, has experienced a substantial rise in robotic implementation,
driven by technological advances, lower costs, and growing consumer demand.
This rapid growth has compelled regulatory bodies, including the Occupational Safety and Health Administration (OSHA), to explore measures for securing safe robot operations as automation progresses~\cite{c0}.
A key safety challenge lies in enabling autonomous mobile robots (AMRs) to respond effectively to irregular situations, such as dropped packages or mechanical breakdowns due to prolonged use.
In practice, addressing these incidents often involve dispatching human workers to clear obstacles, while simultaneously shutting down the robots.
However, halting all robots hinders productivity, highlighting the need for automation solutions that uphold rigorous safety standards while minimizing human intervention and reducing operational downtime.

Early efforts to guarantee safety often relied on physical barriers and fences, such as ``work cells'' in which robots function independently of open-humans~\cite{c1, c6}.
While these measures effectively reduce accidents,
work cells not only require substantial cost of installation,
but also occupy valuable floor space that could otherwise accommodate more production lines or conveyor systems. 
Other common approach is to use
LiDAR or other depth sensors to detect nearby obstacles and adjust their motion accordingly. 
Many such systems rely on potential field methods~\cite{1087247},
in which obstacles exert repulsive forces that increase as they get closer. 
However, when multiple obstacles are present, potential field methods can lead to undesirable behaviors, such as oscillations caused by roughly equal forces from objects on the opposite side. 
In some cases, the robot may even become stuck 
if the sum of surrounding forces cancels out, resulting in controller confusion and a deadlock situation.

\begin{figure}
    \centering
    \includegraphics[width=\columnwidth]{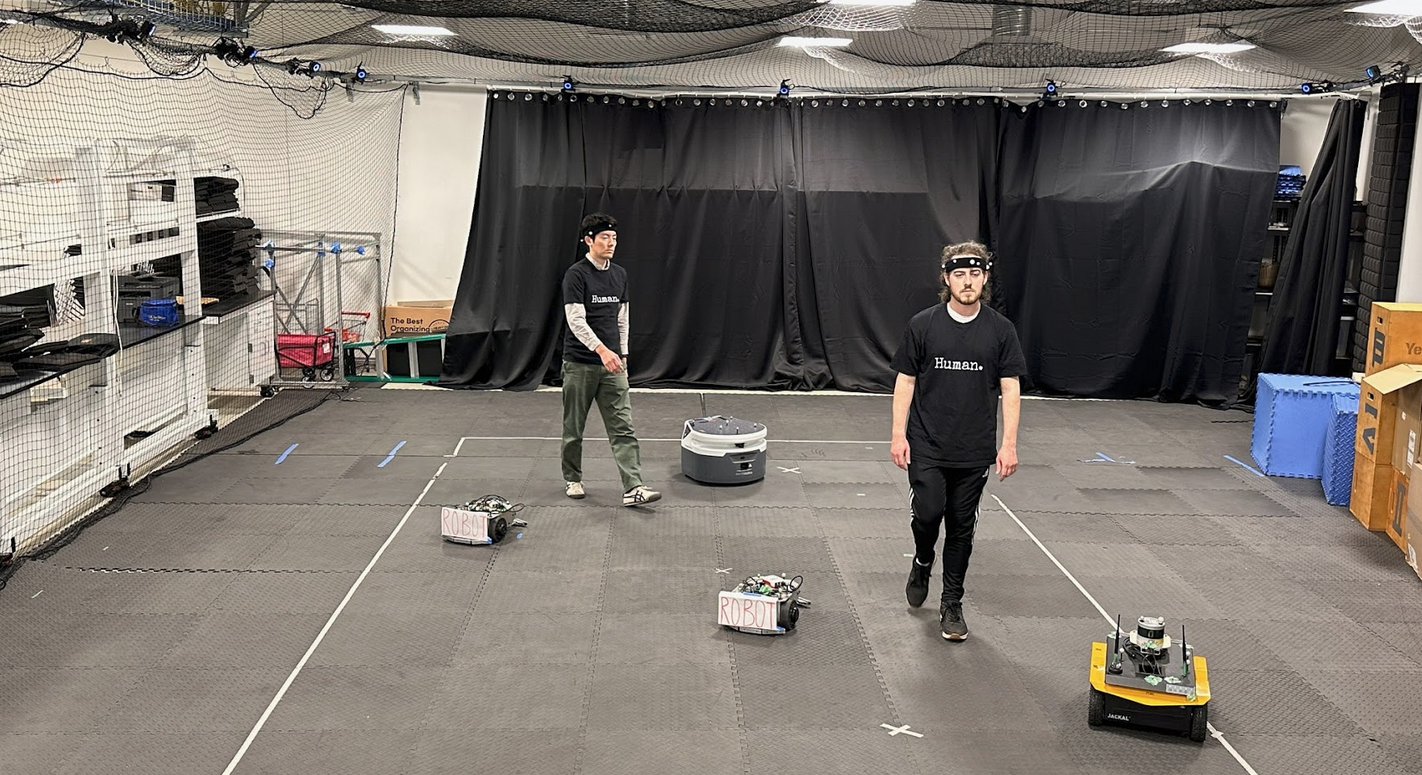}
    \caption{\small Humans walk around the test area, while the robots attempt to reach their goal points without collisions. OpenRMF is used to schedule the high-level task assignment while the on-board Control Barrier Function (CBF) outputs control commands which guide the robot toward safe states.}
    \label{fig:human_robot}
\end{figure}

A promising alternative is the use of Control barrier functions (CBFs) \cite{c9}\cite{c10}\cite{c11}\cite{c26}\cite{c27}\cite{yu2023sequential}. 
By constraining the robot's control inputs, CBFs help ensure that it never enters the hazardous region,
provided the system's dynamics and obstacle positions.
Compared to potential field-based methods, which often become overly conservative to avoid collision,
CBFs invoke conservative controls only if they are necessary to prevent the system from entering unsafe regions.
This inherent nature 
avoids deadlocks when multiple objects surround the robot.
As a result, warehouse operations can continue without unnecessary pauses,
preventing any negative impact on overall productivity.


In this paper, we present an approach that uses data-driven CBFs to enable safe navigation in warehouse environments. 
We train multiple CBFs for obstacle avoidance and multi-robot interactions, 
and then integrate these models as a safety-enhanced local planner within the Open Robotics Middleware Framework (OpenRMF)~\cite{c22},
a platform that implements warehouse logistics and orchestrates global task allocation.
We outline the end-to-end process in detail, from real-world data collection and training sets construction to neural model learning and system integration.
We demonstrate our methodology through comprehensive tests involving static and dynamic obstacles, as well as various robot platforms.
Leveraging the compositional nature of CBFs, our approach can concurrently avoid multiple obstacles, thereby enhancing both safety and efficiency in warehouse operations. The results from our experiments indicate that our method allows the AMRs to operate near their max speed while maintaining a safe distance from obstacles. These results are compiled across a number of scenarios and robot platforms to showcase the generalizability of our method.
In summary, our contributions in this paper are as follows:
\begin{itemize}
\item We employ NCBF-BC \cite{yu25:cbf} to construct a neural representation of control barrier functions.
\item We develop a process for applying learned NCBFs to physical systems, covering data collection, learning, and control derivation (Section \ref{sec:methodology}).
\item We integrate our NCBF into OpenRMF, a high-level task planner (Section \ref{sec:system_overview}).
\item We evaluate our method through experiments with various robot platforms, speeds, and obstacles (Section \ref{sec:EXPERIMENTS}).
\end{itemize}



\section{RELATED WORKS}
Although the introduction of robotics in warehouses is not new, the methods employed to ensure safety are continuously evolving due to technological advancements and governmental policy. Papers such as \cite{c1} discuss how the topic of safety for robots within an industrial setting is approached. They discuss how safety for different types of robots (AGVs, AMRs, manipulators) are approached by the industry. A section is dedicated to the coverage of safety standards for autonomous systems, specifically mentioning laws, ISO standards, IEC standards, as well as general guidelines. A key takeaway from the paper is the need for quicker adoption of safety standards, which are quickly falling behind the pace of technological development and automation adoption. 

Another paper \cite{c5} takes a more theoretical approach in which they evaluate the current trends of industry and hypothesize several future scenarios in which mobile robotic platforms could be applied to factory environments. These hypotheses are then pitched to representatives of German companies, who provide feedback on their feasibility in the short and long term.

The authors in \cite{c7} identify that the logistics of moving material from one place to another within a warehouse is a large source of inefficiency for human workers. This is a prime role for automation via AMRs, by replacing humans in this highly repetitive task. The paper goes on to cover common sensors, localization approaches, planning, management of fleets of robots, and more. They conclude by claiming that AMRs have already made a huge contribution to the industrial warehouse environments, and we can expect to see much more of this as robots become ever cheaper and better performing.

The method within \cite{c6} describes a system for a collaborative human-robot warehouse environment. The authors mention that the classical approach to ensuring safety when it comes to autonomous systems is to isolate them from the human workers via physical barriers or other means, as mentioned earlier. However, the system proposed utilizes dynamic safety fields around the robots of varying danger levels. The three levels include green (clear), yellow (warning), and red (critical) which govern how the robot should react given its surroundings and the potential for collision or other danger. 

Our system operates similarly in that we utilize sensors to determine whether an obstacle is present in the proximity of the robot, however the primary means of safety differs in that we utilize a learned control barrier to generate safe control commands for the current robot platform rather than setting an explicit state which determines the robot behavior. Additionally, one of the key factors in utilizing the control barrier approach is that it can be mathematically shown that the system should never enter an unsafe state given the proper system dynamics and obstacle information \cite{c15}. 

\section{Preliminary}

Consider a dynamical system defined by the vector field $f:\mathcal{X} \times \mathcal{U} \rightarrow \mathcal{X}$ where $\mathcal{X}$ is the state space and $\mathcal{U}$ is the control space.
Let $\mathcal{X}_{u} \subset \mathcal{X}$ denote the unsafe region in which the system's safety constraints are violated.
Control barrier functions are scalar-valued functions over the state space whose spatial gradients can be used to derive controls that keep the system from entering unsafe regions.
\begin{definition}[Control Barrier Functions~\cite{c23}] \label{def:barrier_func}
Let $B:$ $\mathcal{X}$ $\rightarrow$ $\mathbb{R}$ be a continuously differentiable function.
If the zero-superlevel set of $B$, i.e. $\mathcal{C}=\{x\in \mathcal{X}: B(x)\geq 0\}$, is disjoint from the unsafe region of the system, i.e. $\mathcal{C} \cap \mathcal{X}_{u} = \emptyset$.
Moreover, if for any safe state $x \in \mathcal{C}$ and an extended class-$\mathcal{K}_{\infty}$ function ${\alpha} (\cdot)$~\cite{c24}:
{\small\begin{equation} \label{equ:barrier_function}
\begin{aligned}
\max_{u\in \mathcal{U}} L_{f,u} B(x) \geq -\alpha (B(x)),
\end{aligned}
\end{equation}}
where $L_{f,u} B(x)$ is the Lie derivative of $B$ measuring the change of function value over time along the system dynamics under control $u$.
Then $B$ is a control barrier function, and its zero-superlevel set $\mathcal{C}$ is control invariant.
\end{definition}

In this work, we employ \textit{NCBF-BC}~\cite{yu25:cbf} to construct neural representations of control barrier functions from data. The approach describes a process for learning CBFs \cite{c13}\cite{c14}\cite{c16}\cite{c17}\cite{c18} in order to improve the safety of physical robot systems.
NCBF-BC is designed for an offline setting, where no interaction with the system is allowed, and training relies solely on a pre-collected, sparsely labeled dataset.
To make use of unlabeled data with uncertain safety, NCBF-BC uses out-of-distribution (OOD) analysis~\cite{yang2024generalized}, assigning labels based on whether there exist controls capable of driving the system to states that are both \textit{safe} and \textit{in-distribution}. We define a loss function which is used to optimize the parameters of the network. This loss function penalizes the network to enforce positive values with safe states and negative values with unsafe states. Additionally, we include a dynamic feasibility term, which considers the dynamics of the system to promote the system to move towards and stay within safe states.



\section{Methodology - Safety} \label{sec:methodology}


In this section, we present our approach for constructing training datasets from real-world robots and pedestrian trajectories and  leveraging them to learn data-driven neural CBFs to build low-level controllers for collision avoidance.

\subsection{Data Collection}

We begin by describing our two-part dataset, collected from real-world systems, which serves as the basis for learning neural CBFs. 
The first (AMR) portion comprises several seven-dimensional trajectories, each entry specifying the robot’s planar position, yaw angle, linear and angular velocities, and control inputs for target velocities. 
We adopt a discrete control setting (Table \ref{tab:key_control})
in this work, which is sufficient for the desired avoidance behaviors.
The pedestrian portion is two-dimensional, capturing only the pedestrian’s planar coordinates. Both datasets were recorded at 0.1-second intervals using the OptiTrack Motion Capture System~\cite{c19}\cite{c25} for localizing robots and pedestrians. 

We employ a teleoperated control system to manually navigate the robot while pedestrians move naturally at varying velocities. For our experiments, we collect approximately 10 minutes of data for each type of robot and 30 minutes of data from two pedestrians walking at varying velocities to ensure diverse and effective training. Additionally, datasets collected at higher maximum velocities inherently include trajectories at lower velocities, ensuring a comprehensive range of motion dynamics for training.
\begin{table}[t!]
\centering
\small 
\setlength{\tabcolsep}{4pt} 
\renewcommand{\arraystretch}{1.2} 
\begin{tabular}{l c}
    \toprule
    \textbf{Maximum Velocity} & \textbf{Key Control Candidates} \\
    \midrule
    \multirow{2}{*}{0.5 m/s} & \{0.0, 0.3, 0.5\} m/s \\
    & \{-0.8, -0.4, 0.0, 0.4, 0.8\} rad/s \\
    \midrule
    \multirow{2}{*}{1 m/s} & \{0.0, 0.3, 0.5, 1.0\} m/s \\
    & \{-1.2, -0.8, -0.4, 0.0, 0.4, 0.8, 1.2\} rad/s \\
    \midrule
    \multirow{2}{*}{1.5 m/s} & \{0.0, 0.3, 0.5, 1.0, 1.5\} m/s \\
    & \{-1.2, -0.8, -0.4, 0.0, 0.4, 0.8, 1.2\} rad/s \\
    \bottomrule
\end{tabular}
\caption{\small Key Control Candidates. We have three settings with different maximum linear velocity. The first row represents the linear velocity candidates, while the second row represents the angular velocity candidates. The final set of control candidates is obtained by taking the Cartesian product of these two sets.}
\label{tab:key_control}
\end{table}

\subsection{Dynamics Learning}

In this work, we employ multiple robot platforms that, although generally following the differential drive kinematics model, exhibit distinct dynamic characteristics, particularly in maximum linear and angular accelerations.
Because learning neural CBFs requires accurate system dynamics modeling, we begin by training a separate neural dynamics model for each platform.
Note that these models trained on real-world data would account for measurement errors introduced by the localization system.

Denote the robot's planar coordinate on the horizontal and vertical axes by $(x, y)$, respectively, its yaw angle by $\theta$, its linear and angular velocities by $v$ and $\omega$, and its control inputs for desired linear and angular velocities by ($u_{v}$, $u_{\omega}$). 
For one robot platform, We train a neural network $f_{\Theta}: \mathcal{X} \times \mathcal{U} \rightarrow \mathbb{R}^{4}$ to produce bounded refinements over the differential drive kinematics model: 

{\small
\begin{equation}
    \bar{f}_{\Theta}(\underbrace{\begin{bmatrix}
        x \\ y \\ \theta \\ v \\ \omega
    \end{bmatrix}}_{s}, \underbrace{\begin{bmatrix}
        u_{v} \\ u_{\omega}
    \end{bmatrix}}_{u}) = \begin{bmatrix}
        x + \cos{\theta} \cdot (v + \beta \cdot f1_{\Theta}) \cdot dt \\ 
        y + \sin{\theta} \cdot (v + \beta \cdot f1_{\Theta}) \cdot dt \\ 
        \theta + (\omega + \beta \cdot f2_{\Theta}) \cdot dt \\ 
        v + \min(u_{v} - v, m_{v}) + \beta \cdot f3_{\Theta} \\
        \omega + \min(u_{\omega} - \omega, m_{\omega}) + \beta \cdot f4_{\Theta}
    \end{bmatrix},
\end{equation}
}
where $f_{\Theta}(s, u) = \begin{bmatrix}
    f1_{\Theta}, f2_{\Theta}, f3_{\Theta}, f4_{\Theta}
\end{bmatrix}^{T}$ is the model predictions,
$dt$ is the discrete time-step, $\beta$ is the bounding coefficient, 
and $(m_{v}, m_{\omega})$ are the maximal linear and angular accelerations measured from data.
The neural model employs a $tanh$ function in its final layer to ensure bounded outputs.
The parameter $\Theta$ is trained to minimize the mean square error (MSE) between predicted and ground-truth future states.



\subsection{Neural CBF Learning}

Collision avoidance for AMRs can be categorized into three distinct tasks - 
static obstacle avoidance, dynamic obstacle avoidance, and multi-robot interactions. 
For each task, we construct a dedicated training set from the real-world datasets described in Section~\ref{sec:training_preparation}, and train a separate CBF model accordingly. The training set contains labeled safe and unsafe data, as well as the unlabeled data that has uncertain safety.

\subsubsection{Training Preparation}\label{sec:training_preparation}

For \textit{\textbf{static obstacle avoidance}}, we use a five-dimensional input space for the neural CBF model incorporating the robot's relative position with respect to an obstacle, its yaw angle and its velocities. 
To construct the training sets, we proceed as follows.
For each AMR trajectory in the dataset, we randomly sample a coordinate within the map and interpret it as an imaginary static obstacle located along that trajectory.
We introduce a distance threshold $d$ to define collision. 
If the robot remains at least $d$ units away from the obstacle at every time step, the entire trajectory is collision-free, and the corresponding states are labeled as \textit{safe}. 
Otherwise, any states at which the distance is less than $d$ is labeled \textit{unsafe}, and we treat the segment of the trajectory immediately preceding the first unsafe state (of length $\tau$) as \textit{unlabeled}.
Intuitively, when the robot goes from a safe distance to eventually colliding with the obstacle, it must have crossed the optimal safety boundary without taking avoidance actions.
We will follow the annotation procedures of NCBF-BC to recover this optimal boundary.

For \textbf{\textit{dynamic obstacle avoidance}}, the CBF model employs a nine-dimensional input space,
which includes the robot's yaw angle, its velocities, and the pedestrian's relative positions to the robot over the past three ttime steps
Training data preparation follows the same methodology as in the static case,
except instead of randomly sampling obstacle positions, 
we use recorded pedestrian trajectories.
A collision is defined whenever the AMR and pedestrian trajectories intersect within the unsafe distance $d$.

\begin{figure*}[ht]
    \centering
    \begin{subfigure}{0.33\textwidth}
        \centering
        \includegraphics[width=\textwidth]{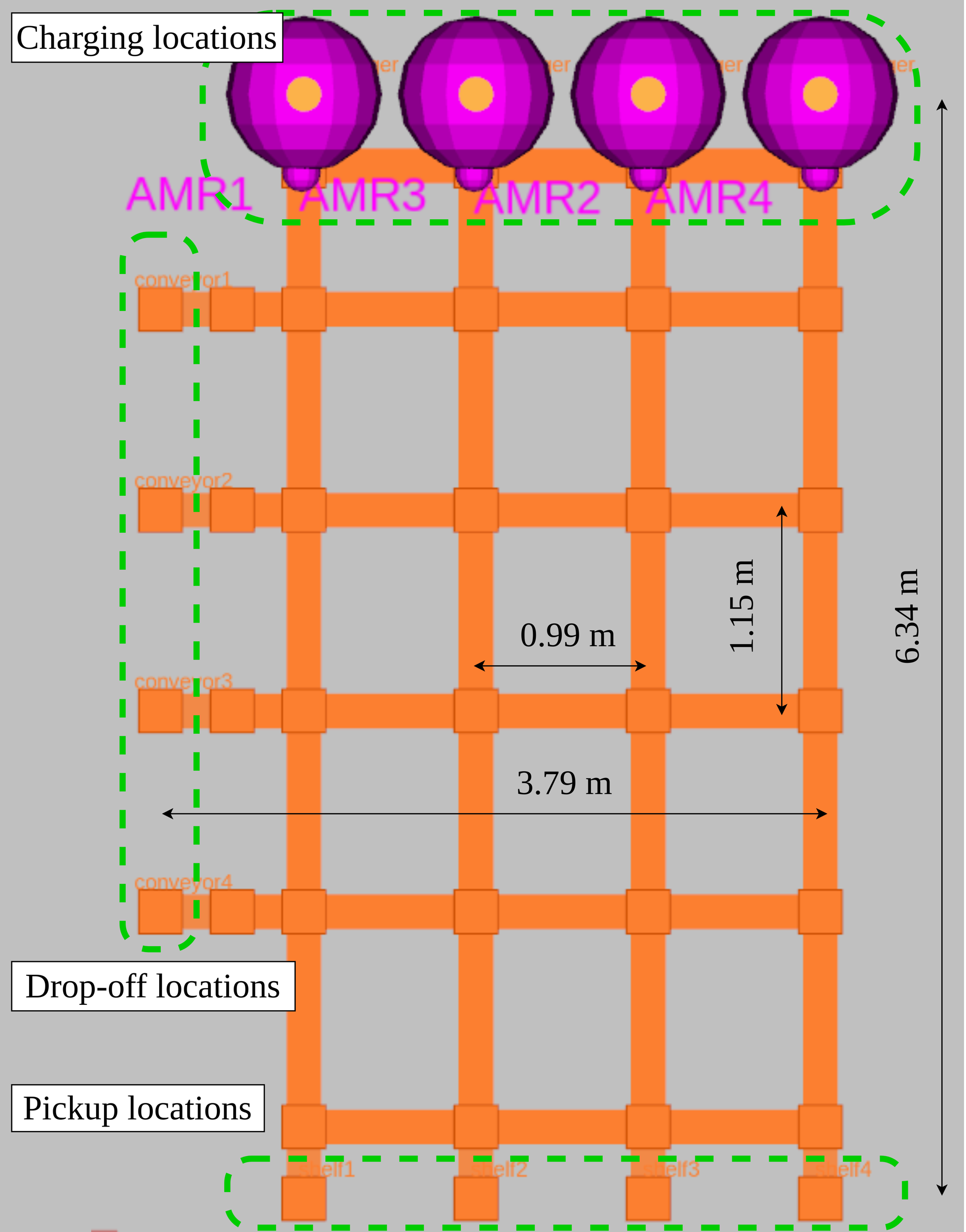}
        \caption{\small Visualization of OpenRMF Map.}
    \end{subfigure}
    \hspace{0.5cm}
    \begin{subfigure}{0.48\textwidth}
        \centering
        \includegraphics[width=\textwidth]{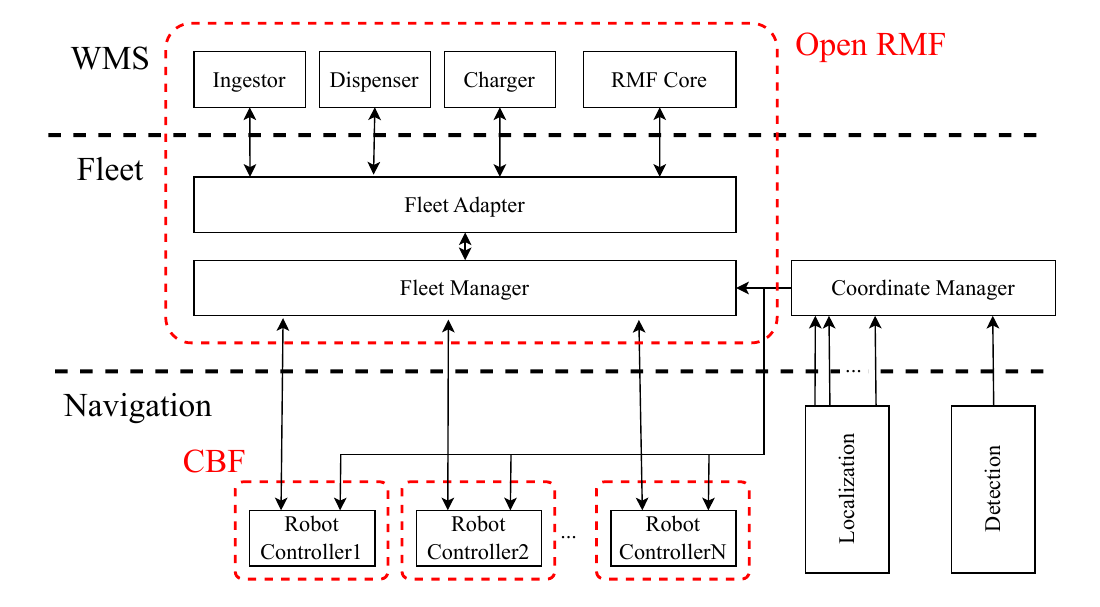}
        \caption{\small System Architecture}
        \vspace{0.4cm} 
        \includegraphics[width=\textwidth]{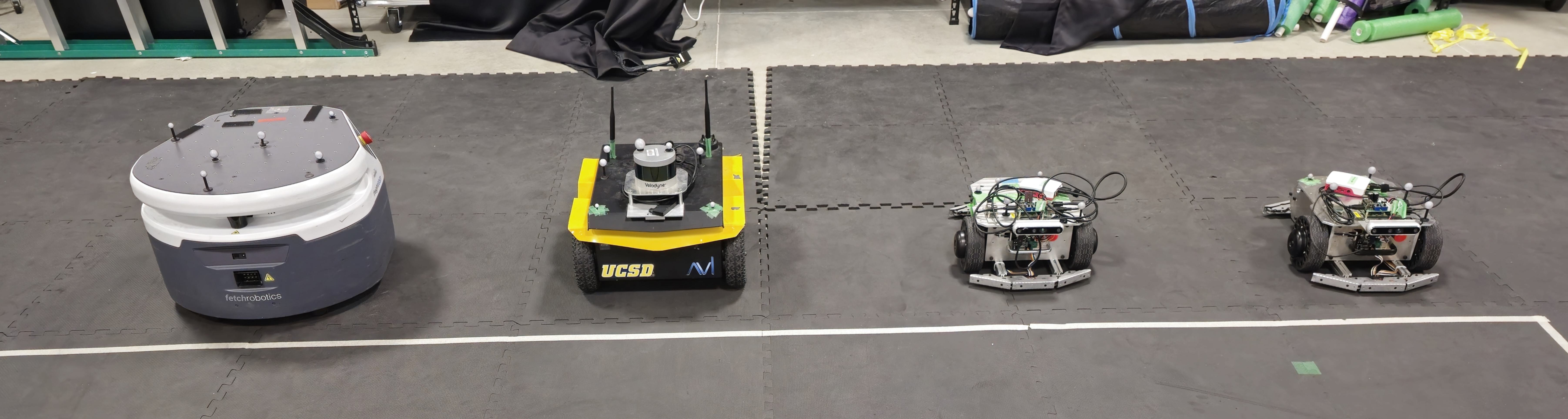}
        \caption{\small Robot Platforms}
    \end{subfigure}
    \caption{\small Overview of the warehouse system: (a) Visualization of OpenRMF Map. The robots are not confined to the illustrated lanes and can freely navigate throughout the environment as needed. (b) System architecture. (c) Real-world robots, from left to right: Freight, Jackal, and two Megarovers.}
    \label{fig:warehouse_system}
\end{figure*}

To ensure safe \textbf{\textit{multi-robot control}}, we adopt a decentralized framework in which each AMR treats the others as uncontrollable.
Let $A1$ be the primary AMR under control, and $A2$ be another AMR.
The CBF model's input space is eight-dimensional, including $A1$'s position relative to $A2$, along with both robots' the yaw and velocity information. 
When additional AMRs are present, we evaluate the CBF pairwise for each.
This approach scales effectively and remains suitably conservative.
In preparing the training sets, we sample pairs of AMR trajectories and consider a collision to occur if the two trajectories intersect within $d$.

\subsubsection{Training Objectives}

Following the approach in NCBF-BC, we learn neural CBFs from the prepared training sets,
while actively annotating the unlabeled states to make full use of the offline data.
To ensure that unlabeled states receive accurate labels, we train an out-of-distribution (OOD) model $R_{\Phi}: \mathcal{X} \rightarrow \mathbb{R}^{2}$.
This model produces two-dimensional scores indicating
how likely it is that a given state belongs to the distribution of previously labeled data.
Concretely, a state $x \in \mathcal{X}$ is considered in-distribution if the two scores satisfy a rejection threshold $c$, namely $R_{\Phi_{1}}(x) > c$ and $R_{\Phi_{2}}(x) > 1 - c$, where $R_{\Phi_{i}}(x)$ is the $i^{th}$ output score.
For any unlabeled state $o$, we assign a safe label if and only if there exists a control action driving $o$ to a future state that is both safe and in-distribution.
This mechanism helps to prevent incorrect labeling caused by false model generalization outside the training data distribution.

A key difference from the original NCBF-BC is that we do not train an actor model for maximally safe controls. 
Instead, given the discrete control setting, we manually perform the \textit{max} operator over the Lie derivative of the CBF to satisfy the invariance condition~(\ref{equ:barrier_function}).
For labeled sets $\mathbb{X}_{s}$ and $\mathbb{X}_{u}$, which include the annotated unlabeled data, the CBF model $B_{\theta}: \mathcal{X} \rightarrow \mathbb{R}$ is trained by minimizing the following loss:

{\small\begin{align}
    &L_{\theta}(\mathbb{X}_{s},\mathbb{X}_{u})\label{eqn:barrier_objective2}\\
    &=\Big(\frac{1}{|\mathbb{X}_{s}|} \sum_{\small x \in\mathbb{X}_{s}} [-B_{\theta}(x)]_{+}\Big) + \Big(\frac{1}{|\mathbb{X}_{u}|} \sum_{x \in \mathbb{X}_{u}} [B_{\theta}(x)]_{+}\Big)\nonumber\\
     &+ \frac{1}{|\mathbb{X}_{s}|} \sum_{x \in \mathbb{X}_{s}}  \Big[-L_{f_{\Theta}, a(x)}B_{\theta}(x) -\alpha \bigg(B_{\theta}(x)\bigg)\Big]_{+},\nonumber
\end{align}}

where:
{\small\begin{align}
    a(x) = \arg\max_{u \in \mathcal{U}} &B_{\theta}\Big(f_{\Theta}(x, u)\Big) \cdot \mathbbm{1}\Big(R_{\Phi_{1}}(f_{\Theta}(x, u)) > c\Big) \\
    &\cdot \mathbbm{1}\Big(R_{\Phi_{2}}(f_{\Theta}(x, u)) > 1 - c\Big)\nonumber
\end{align}}
deriving the control that lead to the safest, in-distribution future state, from the discrete control space.

\begin{table*}[t]
\centering
\small
\setlength{\tabcolsep}{4pt} 
\renewcommand{\arraystretch}{1.2} 
\begin{tabular}{c c c c c c c}
    \toprule
    \textbf{Type} & \textbf{Hidden Layer} & \textbf{Rejection Layer} & \textbf{Rejection Threshold $c$} & \textbf{Unlabeled Horizon $\tau$} & \textbf{Unsafe Range $d(m)$} \\
    \midrule
    Static & [32,32] & [32,32] & 0.25 & 5 & 0.7 \\
    Dynamic & [128,128] & [128,128] & 0.1 & 12 & 0.7 \\
    \bottomrule
\end{tabular}
\caption{\small Network Parameters for Static and Dynamic Avoidance}
\label{tab:network_parameters}
\end{table*}

\subsection{Deriving Controls with Neural CBFs} 

We derive safe controls using well-trained neural CBF models in a sampling-based fashion.
Our approach follows a decentralized setting, where each robot applies a shared control function to navigate its environment while independently managing avoidance of other agents, including robots, pedestrians, and static obstacles.

At each timestamp, the robot receives a list containing the position of all surrounding agents over three past timestamps. For a non-robot agent, we compute its mean velocity with respect to a predefined threshold to determine the agent's type (e.g., pedestrian, robot type). 
Once classification is complete, 
we unroll the learned robot dynamics over discrete controls candidates to predict future states. After that, we iterate through the list of surrounding agents, infer the corresponding neural CBF based on the agent type. Any control candidate yielding unsafe CBF values over the predicted future states are discarded. For the remaining candidates, we select the one that maximize the goal-driven score composed of a velocity tracking term and a goal reaching term.
As the desired velocity is preset, the upper-level layer sends only the goal state to the low-level controller.




\begin{table*}[ht]
\centering
\small
\setlength{\tabcolsep}{4pt} 
\renewcommand{\arraystretch}{1.2} 
\begin{tabular}{l c c cc cc cc}
\toprule
\multirow{2}{*}{\textbf{Type}} 
  & \multirow{2}{*}{\textbf{\# Obstacles}} 
  & \multirow{2}{*}{\textbf{Max Speed}}
  & \multicolumn{2}{c}{\textbf{One Robot Controls}} 
  & \multicolumn{2}{c}{\textbf{Two Robot Controls}} 
  & \multicolumn{2}{c}{\textbf{Three Robot Controls}} \\
\cmidrule(lr){4-5}\cmidrule(lr){6-7}\cmidrule(lr){8-9}
& & \textbf{(m/s)}
  & \textbf{Mean Velocity} & \textbf{Distance} 
  & \textbf{Mean Velocity} & \textbf{Distance} 
  & \textbf{Mean Velocity} & \textbf{Distance} \\
& & 
  & \textbf{(m/s)} & \textbf{(m)} 
  & \textbf{(m/s)} & \textbf{(m)} 
  & \textbf{(m/s)} & \textbf{(m)} \\
\midrule
\multirow{3}{*}{Static} 
  & \multirow{3}{*}{2} 
  & 0.5   & 0.49 & 0.88   & 0.46 & 0.83   & 0.46 & 0.77 \\
& & 1.0   & 0.99 & 0.85   & 0.95 & 0.85   & 0.96 & 0.79 \\
& & 1.5   & 1.39 & 0.89   & 1.36 & 0.85   & 1.29 & 0.78 \\
\midrule
\multirow{3}{*}{Dynamic} 
  & \multirow{3}{*}{1} 
  & 0.5   & 0.49 & 0.98   & 0.49 & 0.98   & 0.47 & 0.8 \\
& & 1.0   & 0.96 & 0.9   & 0.93 & 1.01   & 0.91 & 0.84 \\
& & 1.5   & 1.35 & 1.08   & 1.38 & 0.92   & 1.37 & 0.87 \\
\midrule
\multirow{3}{*}{Dynamic} 
  & \multirow{3}{*}{2} 
  & 0.5   & 0.49 & 0.92   & 0.48 & 0.92   & 0.46 & 0.82 \\
& & 1.0   & 0.95 & 1.05   & 0.92 & 0.86   & 0.9 & 0.85 \\
& & 1.5   & 1.37 & 1.14   & 1.37 & 0.89   & 1.32 & 0.83 \\
\bottomrule
\end{tabular}
\caption{\small Experimental Results For Static and Dynamic Avoidance. One robot setup uses Freight, the two robot setup uses Freight and Jackal, and the three robot setup uses Freight, Jackal and Megarover. Each setup is evaluated at multiple maximum speeds, with pedestrian speed increased accordingly for faster robots. }
\label{tab:static_dynamic_experiment_table}
\end{table*}

A key advantage of these low-level controllers is to handle control delays -
i.e. the time interval between issuing a control command and its actual execution.
Suppose the control delay for a robot platform is $h$ seconds.
When unrolling the system dynamics, we simulate an additional $h$ time delay using the controls already planned (but not yet executed).
In other words, we plan the new controls for the step at which they will actually be applied, while using the previously planned controls to maintain consistent dynamics.

\section{SYSTEM OVERVIEW}
\label{sec:system_overview}







OpenRMF is an open-source orchestration platform for AMRs that manages the warehouse logistics in this project.
It offers a pragmatic framework for coordinating multiple-vendor multiple-robot system, ensuring hardware-agnostic interoperability, and delivering graphical tools and reference implementations - features that make it an excellent choice for integrating the proposed data-driven AMR controllers in industrial warehouses.


Figure \ref{fig:warehouse_system} (b) provides a high-level overview of the platform.
The Warehouse Management System (WMS) serves as the central brain, managing all warehouse operations by assigning tasks and monitoring their progress throughout the facility.
The fleet comprises a collection of AMRs,
each managed through a fleet adapter - a standardized interface that enables seamless communications with multi vendor robot systems through the RMF Core, thereby facilitating efficient task allocation and coordinated navigation.
The project employs the Optitrack system for precise localization, measuring the AMRs' planar coordinates and orientations.
An exemplary warehouse map provided to OpenRMF is shown in Figure \ref{fig:warehouse_system} (a).
This map outlines key operational zones in the scenario, including AMR charging stations, drop-off points and pickup locations.
In this scenario, OpenRMF orchestrates \textit{high-level} task allocation for a series of pick-and-place operations, while the proposed data-driven, decentralized \textit{low-level} controllers handle unforeseen challenges ranging from obstacles and pedestrians to complex multi-robot interactions.
In essence, this integrated framework combines efficient high-level orchestration with robust low-level control, resulting in a scalable, safe, and effective warehouse management system.

The advantages of using the proposed neural CBFs for implementing low-level controls are twofold.
First, their integration into OpenRMF is remarkably straightforward.
By leveraging sampling-based control evaluations over compact, well-trained neural models,
these controllers can be packaged as stand-alone modules that seamlessly augment the existing control architecture.
In contrast, deploying standard ROS1 and ROS2 avoidance stacks in a similarly streamlined manner proves challenging,
as their intricate configuration interfaces and numerous dependencies demand significant setup overhead.
Second, the proposed CBF controllers operate via lightweight message communication, requiring only essential tabular information such as the AMRs' coordinates, orientations and velocities.
By avoiding the transmission of heavy sensor data (e.g., lidar or image observations), these controllers provide a more resource-friendly solution for robust low-level control in AMR applications.

\section{EXPERIMENTS}\label{sec:EXPERIMENTS}



In this section, we present real-world experiments carried out to evaluate the system. 
Three distinct robot platforms (Figure \ref{fig:warehouse_system} (c) are employed to demonstrate the adaptability of our approach in varied hardware configurations.
Notably, all low-level control computations using the data-driven neural CBFs run entirely on each AMR's onboard compute resources - no GPUs are utilized. 



For dynamics learning, we employ two-layer ReLU networks with $64$ neurons per hidden layer. 
The measured maximum linear and angular accelerations for the Freight and Jackal platforms are $m_v = 2.15~\mathrm{m/s^2}$ and $m_w = 2.4~\mathrm{rad/s^2}$, respectively, while the Megarover platform exhibits $m_v = 0.6~\mathrm{m/s^2}$ and $m_w = 2.4~\mathrm{rad/s^2}$. 
We set the bounding coefficient $\beta$ to $1$. 
The measured control delay is $0.1$ seconds for the Freight and Jackal platforms, and $0.2$ seconds for the Megarover platform.
For CBF learning, Table~\ref{tab:network_parameters} details the training parameters, with each task handled separately.






\subsection{Unit Task Evaluation}
\begin{figure*}
    \centering
    \begin{subfigure}[b]{0.24\textwidth}
        \centering
        \includegraphics[width=\textwidth,valign=c]{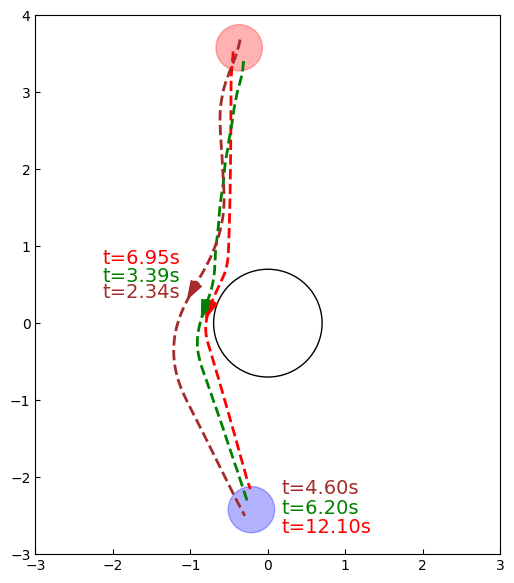}
        \caption{Static Obstacle Analysis}
    \end{subfigure}
    \hspace{2em}
    \begin{subfigure}[b]{0.24\textwidth}
        \centering
        \includegraphics[width=\textwidth,valign=c]{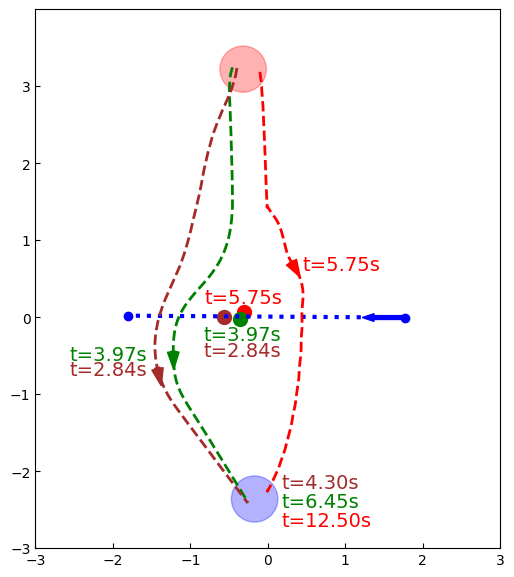}
        \caption{Dynamic Obstacle Analysis}
    \end{subfigure}
    \hspace{2em}
    \begin{subfigure}[b]{0.24\textwidth}
        \centering
        \includegraphics[width=\textwidth,valign=c]{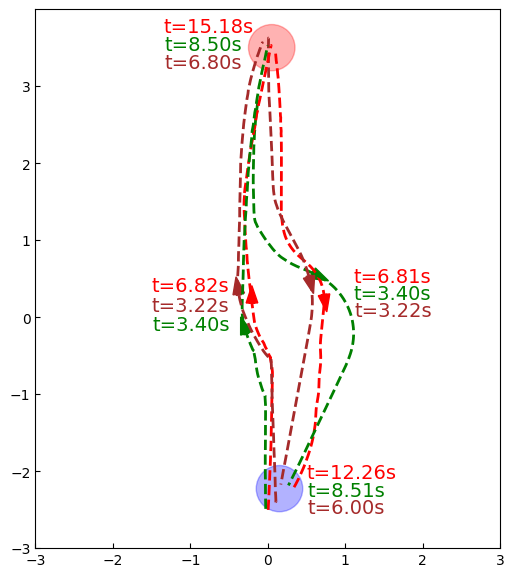}
        \caption{Head-to-Head Analysis}
    \end{subfigure}
    \caption{\small Red, green, and brown trajectories represent robots operating at maximum speeds of 
    0.5\,m/s, 1.0\,m/s, and 1.5\,m/s, respectively. In (a), a circular unsafe range is marked. In (b), the blue trajectory represents a pedestrian’s movement. In (c) shows a head-to-head scenario where the Freight and Jackal start from opposite ends and move toward each other’s initial positions. The timestamps indicate the moments when the robot is at the minimum distance from an obstacle.}
    \label{fig:openrmf_experiments}
\end{figure*}

In our initial experiments, each robot is tasked with navigating from its starting point to a predefined goal,
leveraging the idea that lengthy or complex tasks can be decomposed into smaller unit tasks.
We evaluate performance across various settings that include static and dynamic obstacles, different velocities, multiple robot platforms, and varying robot counts.
Table~\ref{tab:static_dynamic_experiment_table} provides the overall qualitative results, with each setting repeated $10$ times.
Figure \ref{fig:openrmf_experiments} offers visual examples illustrating the distinct behaviors of the trained CBFs under different configurations.
In all experiments, the system successfully reaches its goal without collisions. It maintains its designated speed, and consistently preserves a minimum $0.7$-meter clearance from obstacles, 
adhering to the safety parameters set during training.
At the $1.5$~m/s setting, the system does not reach its maximum speed because the path length is intentionally kept short to remain consistent with the other settings.


As the number of robots and obstacles grows,
higher-speed control candidates are often discarded in favor of safety, leading to speed reductions. 
In the dynamic avoidance scenario, increased velocity prompts more conservative robot behaviors, as indicated by the greater minimum distance in Table \ref{tab:static_dynamic_experiment_table} and illustrated in Figure \ref{fig:openrmf_experiments}b. 
This occurs because the CBF accounts for worst-case situations: faster-moving pedestrians introduce greater uncertainty about their future locations, thus necessitating larger safety margins. 
Conversely, in static avoidance (Figure \ref{fig:openrmf_experiments}a), minimal distance does not necessarily grow with maximum speed. 
This is because the absence of motion uncertainty in static obstacles allows a more consistent avoidance strategy. 
Finally, in a two-robot head-to-head scenario (Figure \ref{fig:openrmf_experiments}c), both robots navigate effectively without being overly conservative.
This outcome demonstrates our method’s ability to balance safety and efficiency in multi-robot contexts.
Since under the proposed decentralized control setting, each robot assumes the other will not react and performs avoidance maneuvers independently, 
they exit the interaction zone quickly while maintaining a safe distance from one another.

\subsection{OpenRMF Experiments}
\begin{table*}
\centering
\setlength{\tabcolsep}{3pt} 
\renewcommand{\arraystretch}{1.2} 
\begin{tabular}{@{}cccccccc@{}}
\toprule
\textbf{\# Pedestrians} & \multicolumn{3}{c}{\textbf{Four Robot Controls (1.0m/s)}} & \multicolumn{3}{c}{\textbf{Two Robot Controls (1.5m/s)}} \\ 
\cmidrule(lr){2-4} \cmidrule(lr){5-7}
                       & \textbf{Minimum Distance} & \textbf{Path Length} & \textbf{Average Velocity} & \textbf{Minimum Distance} & \textbf{Path Length} & \textbf{Average Velocity} \\
                       & \textbf{(m)}      & \textbf{(m)}        & \textbf{(m/s)}  & \textbf{(m)}      & \textbf{(m)}        & \textbf{(m/s)} \\
\midrule
0  & 0.76 & 17.9  & 0.66 & 0.93 & 15.67 & 1.07 \\
1  & 0.72 & 18.73 & 0.65 & 0.83 & 16.78 & 1.07 \\
2  & 0.71 & 19.53 & 0.63 & 0.74 & 18.62 & 1.03 \\
\bottomrule
\end{tabular}
\caption{\small OpenRMF Experiment Results Table. These values exclude periods when the robot is stationary.}
\label{tab:robot_pedestrian_results}
\end{table*}

In the final set of experiments, we use the OpenRMF framework to construct a sample pick-and-place scenario that closely mirrors real-world warehouse operations.
Unlike our earlier experiments, this setup introduces added complexity through a combination of structured tasks.
Specifically, each robot begins at a designated charging station and travels to a pickup location to collect a hypothetical package, then proceeds to a drop-off site before returning to its charging station. 
During these operations, pedestrians move across the warehouse, forcing the robots to perform real-time avoidance maneuvers. 
We summarize the results of these trials in Table \ref{tab:robot_pedestrian_results}. 
To focus on active robot motion, we exclude periods when the robots are deliberately held stationary by OpenRMF’s high-level planner,
where they must stop at the selected junctions
to ensure more organized movement.

When the complexity of the experiment increases through the addition of more robots and more intricate tasks,
our system continues to perform reliably. 
For instance, the minimum distance remains above the unsafe threshold $0.7$ meters.
However, the average velocity does not reach the maximum possible speed.
Because the robots must stop more frequently due to the structured tasks,
the acceleration and deceleration phrases further reduce their overall velocity.
Despite these factors, the results remain consistent with previous unit task experiments.
Notably, we show that the total path length stays within an acceptable range compared to trials without pedestrians,
indicating that only minimal adjustments are required to maintain safety.
By leveraging data-driven CBFs,
our method systematically discards unsafe control candidates while preserving near-maximal velocities whenever conditions allow. 
This strategy ensures that the robots avoid collisions in crowded or dynamic environments without sacrificing overall efficiency.

\section{CONCLUSION}
We have provided an approach which integrates learning-based control barrier functions (CBFs) within the Open Robotics Middleware Framework (OpenRMF) to enable enhanced safety for navigation in multi-robot, multi-agent warehouse environments. Our approach leverages neural CBFs trained on real sensor data to account for the unique dynamics of different robot platforms, including the Freight, Jackal, and Megarover. By incorporating local obstacle avoidance, both static and dynamic, into a high-level fleet management system, we achieve collision-free coordination without imposing significant computational overhead.

Additionally, we conducted real-world experiments to showcase the effectiveness of our method across varying speeds, numbers of obstacles, and pedestrian interactions. In nearly all scenarios, the robots respected a minimum safe distance to obstacles or other agents, even under tight space constraints and with multiple concurrent tasks. While the system remains conservative when the environment is crowded or at higher speeds, it consistently prioritized safety over task completion time.

\addtolength{\textheight}{-22cm}   
\bibliographystyle{IEEEtran}
\bibliography{newlib}

\end{document}